\documentclass[5p,number,sort&compress,times,fleqn]{elsarticle}

\pdfoutput=1

\usepackage[utf8]{inputenc} 

\usepackage{natbib}
\usepackage[hidelinks]{hyperref}
\usepackage{units}
\usepackage{nicefrac}
\usepackage{amsmath}
\usepackage{amssymb}
\usepackage{lipsum}
\usepackage[textsize=footnotesize]{todonotes}

\usepackage{array}
\usepackage{booktabs}
\usepackage{multirow}
\usepackage{threeparttable}

\newlength{\autocolumnwidth}
\newlength{\firstcolumnwidth}
\newcount\nColumns

\newlength{\plotwidth}

\usepackage{tabularx}
\newcolumntype{Y}{>{\centering\arraybackslash}X}

\usepackage{siunitx}
\DeclareSIUnit\px{pixels}
\usepackage{float}
\usepackage{glossaries}
\glsdisablehyper

\newacronym{ANN}{ANN}{artificial neural network}
\newacronym{PSD}{PSD}{particle size distribution}
\newacronym{TEM}{TEM}{transmission electron microscopy}
\newacronym{WST}{WST}{watershed transformation}
\newacronym{UE}{UE}{ultimate erosion}
\newacronym{HT}{HT}{Hough transformation}
\newacronym{GMD}{GMD}{geometric mean diameter}
\newacronym{GSD}{GSD}{geometric standard deviation}
\newacronym{MSE}{MSE}{mean squared error}
\newacronym{CE}{CE}{cross-entropy}

\usepackage[babel]{csquotes}

\usepackage[final]{changes}

\definechangesauthor[name={reviewer 1}, color=red]{R1}
\definechangesauthor[name={reviewer 2}, color=yellow]{R2}
\definechangesauthor[name={reviewer 3}, color=blue]{R3}
\definechangesauthor[name={author}, color=green]{GK}

\newcommand{\subfiguretitle}[1]{{\vspace{0.5\baselineskip}\mbox{\large\fontfamily{phv}\fontseries{b}\selectfont\smash{\centering#1}}}\\\vspace{0.2em}}

\journal{Powder Technology}

\begin{document}
	
	\setlength{\columnwidth}{252pt}
	\setlength{\plotwidth}{0.97\columnwidth}

\begin{frontmatter}

\title{Fully automated primary particle size analysis of agglomerates on transmission electron microscopy images via artificial neural networks}

\author{M. Frei\corref{mycorrespondingauthor}}
\cortext[mycorrespondingauthor]{Corresponding author. Tel.: +49 203 379--3621}
\ead{max.frei@uni-due.de}

\author{F.E. Kruis\corref{}}

\address{Institute of Technology for Nanostructures (NST) and Center for Nanointegration Duisburg-Essen (CENIDE)\\University of Duisburg-Essen, Duisburg, D-47057, Germany}

\begin{abstract}
There is a high demand for fully automated methods for the analysis of primary particle size distributions of agglomerates on transmission electron microscopy images. Therefore, a novel method, based on the utilization of artificial neural networks, was proposed, implemented and validated.

The training of the artificial neural networks requires large quantities (up to several hundreds of thousands) of transmission electron microscopy images of agglomerates consisting of primary particles with known sizes. Since the manual evaluation of such large amounts of transmission electron microscopy images is not feasible, a synthesis of lifelike transmission electron microscopy images as training data was implemented.

The proposed method can compete with state-of-the-art automated imaging particle size methods like the Hough transformation, ultimate erosion and watershed transformation and is in some cases even able to outperform these methods. It is however still outperformed by the manual analysis.

\end{abstract}


\end{frontmatter}


\section{Introduction}
The properties of nanomaterials (e.g. mechanical, optical, catalytic, biological,~etc.) are determined by their characteristic length as well as their shape \cite{Guisbiers.2012}. In case of nanoparticles, the characteristic length is usually some kind of equivalent diameter (e.g. hydrodynamic, volume-based, surface-based, etc.) \cite{Calvert.1990}. There are various techniques available for the determination of the different types of equivalent diameters, one of them being \gls{TEM} \cite{Cho.2013}. 

The determination of \glspl{PSD} of non-agglomerated\footnote{In the context of this work an agglomerate is defined as \blockquote{[...] a collection of weakly bound particles [...]}, according to the definition of the European Commission \cite{EC.2011}.} particles on \gls{TEM} images can already be performed partially or fully automated since several decades \cite{King.1982, King.1984}. By contrast, the determination of \glspl{PSD} of agglomerated particles still is a very challenging task for conventional image processing algorithms \cite{Temmerman.2014} and therefore still depends on the recognition of primary particles by a human operator. This practice is not only laborious, expensive and repetitive but also error-prone due to the subjectivity and exhaustion of the operator \cite{Allen.2003}.

Therefore, a new approach to imaging particle size analysis, with help of \glspl{ANN}, was proposed, implemented and validated. The \glspl{ANN} were trained to determine the projected areas of primary particles of agglomerates via regression and the number of primary particles per agglomerate via classification. A major challenge of this approach is the fact that the training of \glspl{ANN} requires up to several hundreds of thousands of samples with known properties to avoid overfitting, i.e. that the \gls{ANN} is not able to generate correct outputs for unknown samples because it is too specialized on its training data  \cite{Babyak.2004}. 

Unfortunately, there is no publicly available source of already evaluated samples and a manual evaluation of so many images is hardly feasible, due to the reasons given before. Therefore, inspired by Kruis et al. \cite{Kruis.1994}, \gls{TEM} images with defined characteristics were synthesized and used for the training of the \glspl{ANN}. To speed up the image synthesis as well as the training of the \glspl{ANN} the necessary calculations were performed on graphics processing units, whenever possible.

\section{State of the Art}
Most of the conventional imaging particle size analysis methods are enhancements or combinations of the \gls{WST} \cite{Jung.2010,Shu.2013,Cheng.2009}, \gls{UE} \cite{Wang.2016,Park.2013} and/or \gls{HT} \cite{Ballard.1981,Merlin.1975,Xu.1990,Kruis.1994}. Usually, these methods have one or more parameters, which can be used to control the sensitivity of the primary particle detection. While the availability of such parameters enhances the versatility of the methods, it also impedes their adaptability, because the parameters need to be determined manually prior to the image analysis. Apart from that, none of the methods can compete with the accuracy of a manual evaluation.

So far, to the best of our knowledge, there have only been very few publications concerning the image-based measurement of \glspl{PSD} with help of \glspl{ANN}, all of which had a mining context \cite{Ko.2011,Hamzeloo.2014}. Therefore, regarding the particle sizes, the image coverage and the utilized image detectors, the specimens that were analyzed in these publications are very different from the specimens that were used in the work at hand. In the context of nanoparticles, several \gls{ANN}-based methods for the determination of \glspl{PSD} have been published. However, all of these methods are based on fundamentally different measurement principles (e.g. dynamic light scattering \cite{Stegmayer.2006,Gugliotta.2009}, laser diffraction \cite{Guardani.2002,Nascimento.1997} and focused beam reflectance measurement \cite{Mausse.2006,Irizarry.2017}). 

\section{Artificial Neural Networks}

The next sections will address the basic concepts of feed-forward \glspl{ANN}. The interested reader may refer to literature for a comprehensive treatment of the historical development \cite{Schmidhuber.2015}, the fundamentals \cite{Haykin.2009,Kriesel.2007} and the design \cite{Hagan.2016} of \glspl{ANN}.

\begin{figure}
	\centering
	\includegraphics[width=\linewidth]{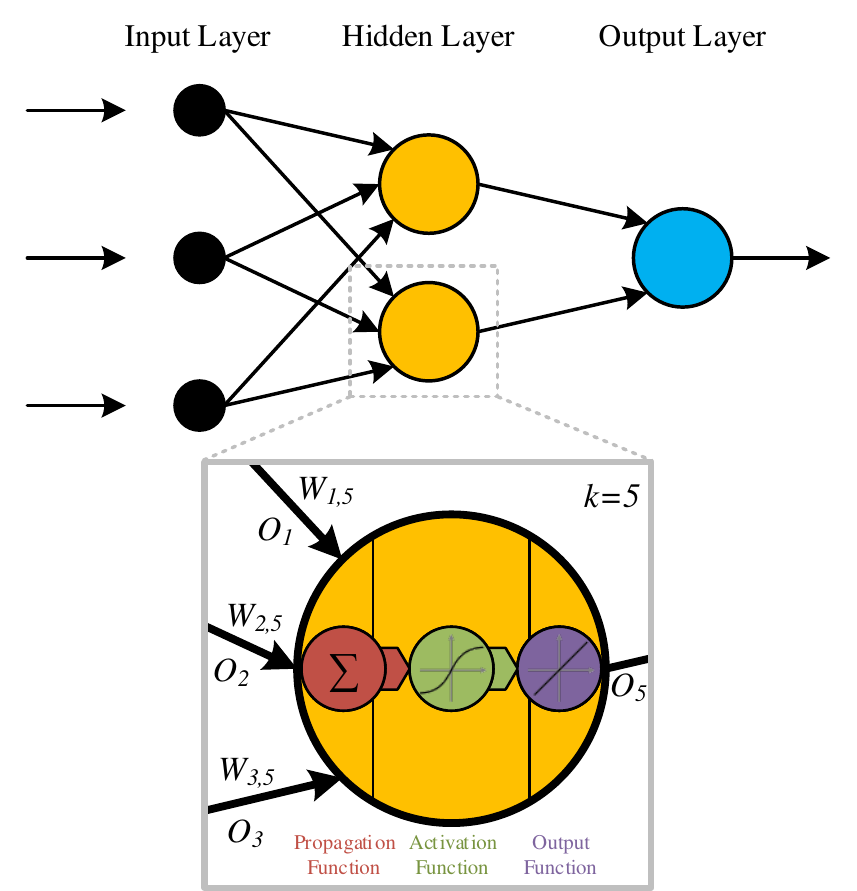}
	\caption{Schematic representation of the common structure of an \gls{ANN}: input layer, hidden layer and output layer. Magnified: Schematic representation of a neuron.}
	\label{fig:NeuronalNet}
\end{figure}

An \gls{ANN} consists of simple computational units called neurons, which are grouped in layers and are unidirectionally connected to each other by weighted connections. In its simplest form an \gls{ANN} has only one input and one output layer but usually, these are separated by one or more hidden layers (Figure \ref{fig:NeuronalNet}). In feed-forward \glspl{ANN} data may only flow from left to right and there are no loops within the \gls{ANN}. \cite{Kriesel.2007}

A neuron typically contains three consecutive functions,\linebreak which process the data before it is used as input for the following neurons (Figure \ref{fig:NeuronalNet}):
\begin{enumerate}
	\item The propagation function, which transforms the outputs of previous neurons into an input for the current neuron, based on the individual weights of the connections.
	\item The activation function, which computes the activation (also referred to as excitement) of the current neuron, based on the input.
	\item The output function, which transforms the activation into the output of the current neuron.
\end{enumerate}

\subsection{Propagation Function}
\label{sec:Theory-PropagationFunction}

Usually, a neuron $k$ is connected to multiple preceding neurons $j_1,j_2,...,j_N$ with the corresponding outputs $\hat{O}=(O_1,O_2,...,\allowbreak O_N)$ via weighted connections with the associated weights $\hat{W}_k=(W_{1,k},W_{2,k},...,W_{N,k})$. Since the activation function $f_{act}$ usually expects a scalar input, it is necessary to transfer the vectorial output $\hat{O}$ of the preceding neurons into a scalar input $I_k$ for the activation function of the current neuron $k$ by means of a propagation function $f_{prop}$. The most common choice for the propagation function is the weighted sum: \cite{Kriesel.2007} 
\begin{equation*}
\label{eq:PropagationFunction}
I_k = f_{prop}(\hat{W}_k,\hat{O}) = \sum_{i=1}^{N}{\left(W_{i,k} \cdot O_i\right)} 
\end{equation*}

\subsection{Activation Function}
\label{sec:Theory-ActivationFunction}

The activation $A_k$ of a neuron $k$ is used as an input for its output function $f_{out}$, to compute its output level $O_k$. It depends on the input $I_k$ of the neuron, the activation function $f_{act}$ (also called transfer function) and an individual additive bias term $B_k$ \cite{Haykin.2009}:
\begin{equation*}
\label{eq:ActivationFunction}
A_k = f_{act}(I_k+B_k)
\end{equation*}
The most common activation functions are sigmoid functions (e.g. hyperbolic tangent) and the identity function \cite{Haykin.2009}. For the output layers of classification \glspl{ANN} (see section \ref{sec:Learning}) a very common activation function is the softmax function \cite{Bendersky.2016}.

\subsection{Output Function}
\label{sec:Theory-OutputFunction}

The output function $f_{out}$ is used to calculate the output $O_k$ of a neuron based on its activation \cite{Kriesel.2007}:
\begin{equation*}
\label{eq:OutputFunction}
O_k = f_{out}(A_k)
\end{equation*}

Often, the output function is the identity function, so that the output of the neuron equals its activation. Normally, all neurons of an \gls{ANN} possess the same output function $f_{out}$. \cite{Kriesel.2007}

\subsection{Learning}
\label{sec:Learning}
In the context of \glspl{ANN} learning comprises the gradual modification of the properties of an \gls{ANN}, so that it can correctly predict a certain output for a certain input. The properties of an \gls{ANN} which are changed during the process of learning are usually the weights of its neurons. Therefore, after the completion of the learning process the weights represent the "knowledge" of an \gls{ANN}. \cite{Kriesel.2007}
\newpage
Generally, learning strategies can be grouped into two categories \cite{Kriesel.2007}:
\begin{itemize}
	\item Supervised learning: The \gls{ANN} is supplied with input data. The error made by the \gls{ANN} is determined by comparing its output data with the desired correct output data, called target data and used to improve the weights and thereby the performance of the \gls{ANN}. This process is being repeated until the performance of the \gls{ANN} is sufficiently good or does not improve within a predefined number of epochs, i.e. repetitions.
	\item Unsupervised learning: The \gls{ANN} is only supplied with input data and tries to find similarities and patterns by itself. 
\end{itemize}
In this work, only supervised learning was utilized.
\\

There is a vast variety of different tasks \glspl{ANN} can learn to solve by means of supervised learning, many of which can be categorized into two groups \cite{Goodfellow.2016}:
\begin{itemize}
	\item Classification: When being supplied with a certain input, the \gls{ANN} outputs a discrete class. An example of a classification could be the assignment of a genre (the output) to a song, based on the year it was published, its length and its dynamics (the inputs).
	\item Regression: When confronted with a certain input, the \gls{ANN} outputs a continuous value. An example of a regression could be the determination of the price of a house (the output), based on the year it was built, its location and its number of rooms (the inputs).
\end{itemize}

For this publication, \glspl{ANN} were used for classification as well as regression. Firstly, to classify agglomerates with respect to the number of primary particles they incorporate and secondly to regress the individual areas of the primary particles of an agglomerate.

As mentioned before, a supervised learning process is based on the comparison of the actual outputs $\hat{O} = (O_1,O_2,...,O_N)$ of an \gls{ANN} with the target outputs $\hat{T} = (T_1,T_2,...,T_N)$. To compare the performances of \glspl{ANN} it is necessary to introduce a criterion for their performance. This criterion is usually implemented in form of a cost (also loss) function. The smaller the cost $C$ of an output, the better the performance of the \gls{ANN}. Depending on the learning task (regression or classification) of the \gls{ANN} different cost functions, e.g. mean squared error or cross-entropy (see equation \ref{eq:MSE} and equation \ref{eq:CE} respectively), can be applied. \cite{Hagan.2016} 

For regression \glspl{ANN} the mean squared error is the most common cost function \cite{Hagan.2016,MATLAB.2016}:
\begin{equation}
\label{eq:MSE}
C_{MSE}(\hat{O},\hat{T}) =\frac{1}{N}\sum_{i=1}^{N}(T_i-O_i)^2
\end{equation}

In contrast, the cross-entropy function is usually used for classification \glspl{ANN} \cite{Hagan.2016,MATLAB.2016}:
\begin{equation}
\label{eq:CE}
C_{CE}(\hat{O},\hat{T}) =-\frac{1}{N}\sum_{i=1}^{N}\left(T_i \cdot \ln O_i\right)
\end{equation}

\section{Workflow}

\begin{figure}
	\centering
	\includegraphics[width=0.95\linewidth]{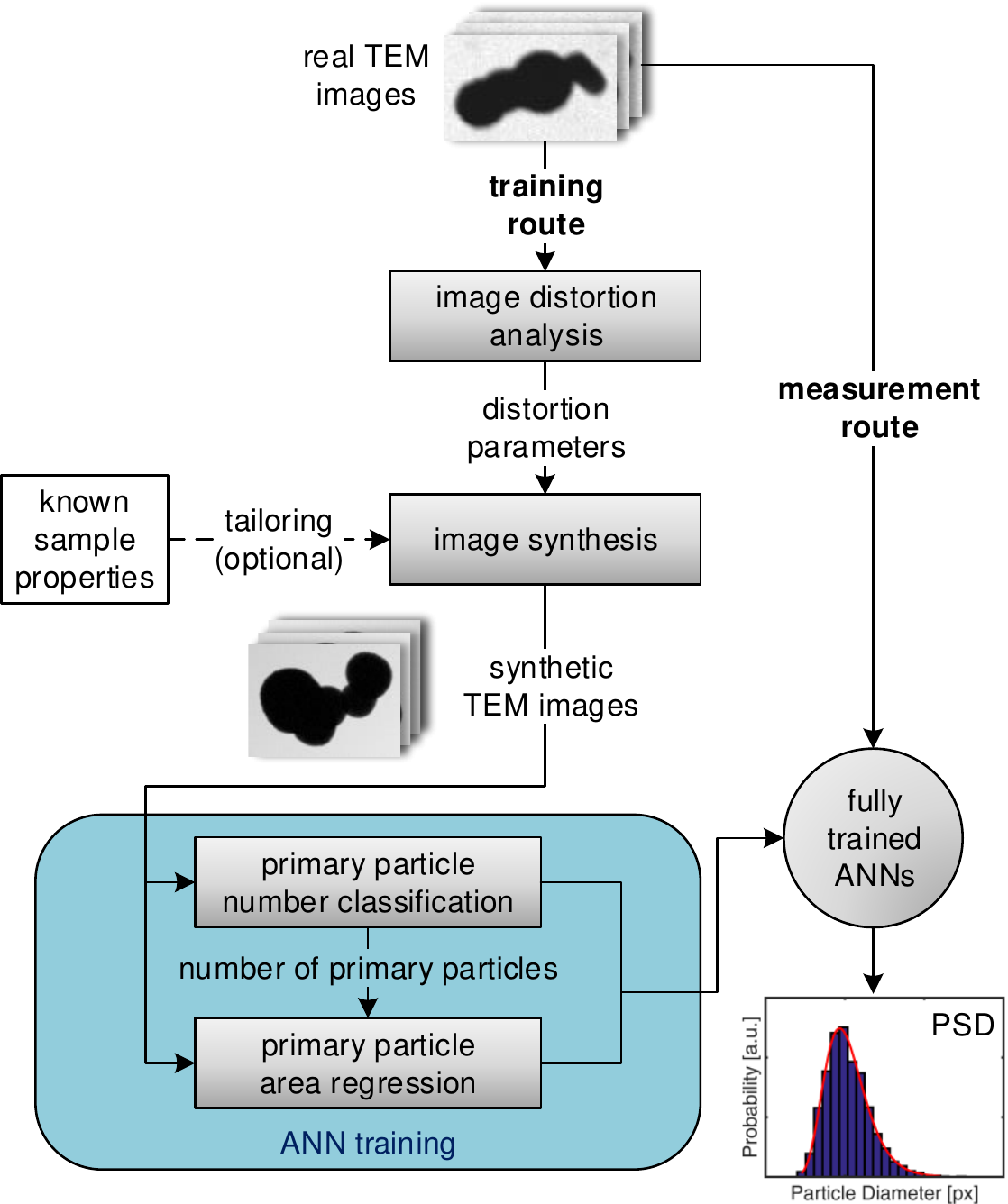}
	\caption[Workflow]{Workflow of the proposed method.}
	\label{fig:Workflow}
\end{figure}

The workflow of the proposed method consists of a training and a measurement route (Figure \ref{fig:Workflow}).

In the first step of the training route, the distributions of the distortion parameters of real \gls{TEM} images are determined, to use them during the image synthesis to obtain more lifelike synthetic \gls{TEM} images. Three characteristic image distortions are considered and quantified with suitable measurands (Table \ref{tab:ImageDistortionMeasurands}).

\nColumns=2
\setlength{\firstcolumnwidth}{2cm}
\setlength{\autocolumnwidth}{(\columnwidth-\tabcolsep*(\nColumns*2)-\firstcolumnwidth)/(\nColumns-1)}
 
\begin{table}
	\centering
	\caption{Measurands of characteristic image distortions.}
	\renewcommand{\arraystretch}{1.2}
	\label{tab:ImageDistortionMeasurands}
	\begin{tabular}{>{\centering\bfseries}m{\firstcolumnwidth} >{\centering\arraybackslash}m{\autocolumnwidth}}
		\toprule
		\toprule
		\textbf{Distortion} & \textbf{Measurands}\\
		\midrule
		blur & Tenenbaum's gradient focus measure \cite{Tenenbaum.1970} \\
		noise & standard deviation of the pixel intensities \\ 
		nonuniform illumination & 1-1-polynomial fit of the pixel intensities \\
		\bottomrule
	\end{tabular}
\end{table}

The image synthesis is a central element of the proposed method because it represents a feasible way to obtain the large number of images with known target properties, which is necessary for the training of the \glspl{ANN}. It is of utmost importance to synthesize images which are as lifelike as possible, to provide realistic training data for the \glspl{ANN}. Elsewise, the \glspl{ANN} might perform well on the training data but will fail once they are applied to real \gls{TEM} images. Additionally, the image synthesis provides an opportunity for the incorporation of a priori knowledge about the product, which is going to be analyzed. Thus, the training data of the \glspl{ANN} can be tailored to the product properties, to facilitate the training and improve the measurement quality.

To reduce the amount of input data and thereby the computational effort during the training and utilization of the \glspl{ANN}, the input images are preprocessed. As preprocessing, 13 features (Table \ref{tab:AgglomerateFeatures}) of each agglomerate are determined and used as input of the \glspl{ANN}. 

There are two different types of \glspl{ANN}, which are used during the determination of primary \glspl{PSD} via the proposed method. On the one hand, there is a classification \gls{ANN}, which determines the number of primary particles an agglomerate incorporates. On the other hand, there are multiple regression \glspl{ANN}, which are utilized for the regression of the primary particle areas of an agglomerate, depending on the number of primary particles it incorporates. This means that for each class of agglomerates an individual regression \gls{ANN} is used. The reason for this two-step strategy is the fact that the number of outputs of an \gls{ANN} needs to be defined a priori and cannot be determined by the \gls{ANN} itself during its run-time. It is important to note that this limits the maximum number of primary particles an agglomerate may contain while still being analyzed, to the number of regression \glspl{ANN} which are trained beforehand. Agglomerates containing a higher number of primary particles are simply sorted out by the classification \gls{ANN} and thereby excluded from the analysis. For this publication, five regression \glspl{ANN} were trained, i.e. the number of primary particles per agglomerate was limited to five.

After the training of the \glspl{ANN} has been finished, they can be used for the determination of primary \glspl{PSD} via the measurement route.

\begin{figure*}
	\includegraphics[width=\textwidth]{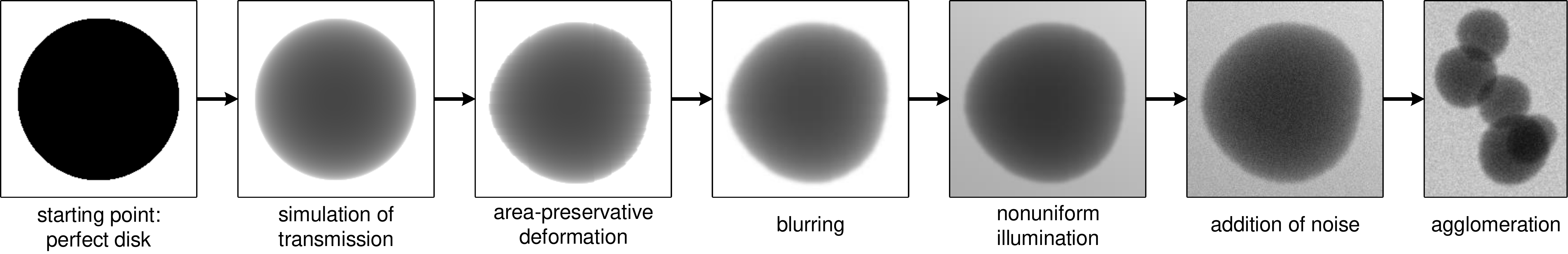}
	\caption{Elementary steps of the image synthesis.}
	\label{fig:ImageSynthesis}
\end{figure*}

\section{Image Synthesis}
Starting with a perfectly circular black disk, the image synthesis basically consists of six steps (see also figure \ref{fig:ImageSynthesis}):

\begin{enumerate}
	\item \textit{Simulation of electron transmission}\\
	For an amorphous spherical object with an outer radius~$R$, the transmission ratio at a radius ${0 \leq r \leq R}$ is given by:
	\begin{equation*}
	\frac{I_T}{I_0} = \exp{(-2 c_T \cdot \sqrt{R^2-r^2} )}
	\end{equation*}
	
	$I_0$ and $I_T$ are the intensities before and after the transmission and $c_T$ is the transmission coefficient, which depends on the density, the atomic weight and number of the material, as well as the aperture angle and the acceleration voltage of the instrument. \cite{Hornbogen.2009}
	
	\item \textit{Area-preservative deformation}\\
	Until now the synthesized particles are perfectly spherical and their two-dimensional projections are therefore perfectly circular. Since this does not apply to most naturally occurring particles, it is necessary to deform the circular form. The challenge of the deformation process is the preservation of the projection area, as it is the measurand of the proposed method.
	\item \textit{Blurring}\\
	The blur synthesis is based on the distribution of the standard deviation of the Gaussian blur, which was determined based on Tenenbaum's gradient focus measure \cite{Tenenbaum.1970}, during the image distortion analysis of real \gls{TEM} images.
	\item \textit{Nonuniform illumination}\\
	For the synthesis of the nonuniform illumination, the distributions of the parameters of the 1-1-polynomial, which were determined during the image distortion analysis of real \gls{TEM} images, are utilized.
	\item \textit{Addition of noise}\\
	As noise for the synthetic \gls{TEM} image, a Gaussian noise with a standard deviation according to the distribution, which was determined during the image distortion analysis of real \gls{TEM} images, is used.
	\item \textit{Agglomeration}\\ 
	So far, the image synthesis has been limited to single particles. However, since the \glspl{ANN} need to be trained on images of agglomerates, an agglomeration algorithm was implemented. Due to the large number of images, which are necessary for the training of the \glspl{ANN}, one of the most important requirements concerning the agglomeration algorithm is a high computation speed. Therefore, complex agglomeration simulations (e.g. kinetic particle simulations) were not an option. Instead, the agglomerates are constructed geometrically, by joining spheres at random surface points, while avoiding overlaps of the spheres.
\end{enumerate}

Ultimately, the image synthesis yields synthetic \gls{TEM} images, which are lifelike to a large extent (see supplementary material).

\section{Primary Particle Number Classification}
The following section will elaborate upon the design, training, validation and results of the \gls{ANN}, which was used for the classification of agglomerates according to their number of primary particles (hereinafter called ParticleNumberNet). Since each class of agglomerates requires its own \gls{ANN} for the regression of its primary particle areas (hereinafter called ParticleAreaNets), it was necessary to limit the number of classes, i.e. the maximum number of primary particles per agglomerate, to reduce the design effort, as well as the computation times during the training and utilization of the \gls{ANN}. It was therefore decided to limit the classification to six classes: one for each number of primary particles from one to five and one for agglomerates containing more than five primary particles, which are not being analyzed.

\subsection{Training, Validation and Test Data}

The input data set which was used for the training of the ParticleNumberNet was generated by synthesizing 100,000 images of each class of agglomerates, where the class of agglomerates which consist of more than five primary particles contained agglomerates with up to 10 primary particles. The area of the primary particles was homogeneously distributed across all synthesized images and varied in the range of approximately \SIrange{500}{6500}{\px}. The homogeneous distribution of the primary particle area was important to prevent the \gls{ANN} from developing preferences concerning the recognition of particles with areas overrepresented in the input data.

Subsequently, the images were segmented, i.e. the foreground was separated from the background, and the agglomerates were characterized by 13 features (Table \ref{tab:AgglomerateFeatures}). To prevent features with large ranges (e.g. area) from overriding features with small ranges (e.g. mean intensity), the features were normalized based on their individual ranges.

The target data for the training of the ParticleNumberNet consisted of a total of 600,000 binary vectors with six components each, where each of the six components represented the probability to belong to one of the six classes, i.e. either 0 or~1. The input and target data was divided into three subsets (Table \ref{tab:ParticleNumber_DataDivision}), to evaluate the generalization abilities of the \gls{ANN} and to allow the utilization of the early stopping method \cite{Goodfellow.2016}.

\subsection{Structural Optimization}
\label{sec:PrimaryParticleNumberClassification-StructuralOptimization}

The final structure of the ParticleNumberNet and the learning strategies utilized during its training are given in tables \ref{tab:ParticleNumber_Structure} and \ref{tab:ParticleNumber_Learning} respectively. 

To yield optimal results, the structure of the ParticleNumberNet was optimized iteratively. Some structural features of \glspl{ANN} depend solely on the structure of the input and the target data. Every \gls{ANN} has exactly one input layer, whose number of neurons $N_i$ is equal to the number of components of an input sample (in case of the ParticleNumberNet $N_i=13$). Additionally, every \gls{ANN} has exactly one output layer, whose number of neurons is equal to the number of components of a target sample (in case of the ParticleNumberNet $N_o=6$). In contrast, the number of hidden layers and the numbers of neurons $N_{h,i}$ in those layers can be modified freely. However, it is uncommon for feed-forward \glspl{ANN} to have more than two hidden layers \cite{Hagan.2016}, because two hidden layers already allow the \gls{ANN} to approximate all possible kinds of functions \cite{Heaton.2008}. Therefore, only \gls{ANN} structures with up to two hidden layers were tested as candidates for the ParticleNumberNet and the ParticleAreaNets.

More important than the number of hidden layers is the number of neurons within these layers. Unfortunately, there are no strict rules for the dimensioning of hidden layers. \cite{Heaton.2008} 

However, there are rules of thumb which can be used as an orientation \cite{Heaton.2008}:

\vspace{-\abovedisplayskip}
\noindent\begin{tabularx}{\columnwidth}{@{}c@{}X@{}c@{}X@{}c@{}} 
	 &
	{\setlength{\mathindent}{0cm}
		\begin{align} 
		\label{eq:RuleOfThumb1}
		&&N_o &\leq N_{h,i} < 2N_i \\
		&\Rightarrow &6 &\leq N_{h,i} < 26 \nonumber
		\end{align}} &
	\hspace{0cm} &
	{\setlength{\mathindent}{0cm}
		\begin{align} 
		\label{eq:RuleOfThumb2}
		&&N_{h,i} &\approx \nicefrac{2}{3}N_i+N_o \\ 
		&\Rightarrow &N_{h,i} &\approx 15 \nonumber
		\end{align}} &
\end{tabularx}
\vspace{-2\belowdisplayskip}

To find the optimum for the number of neurons of the two hidden layers, all possible combinations of ${1 \leq N_{h,1} < 26}$ and ${0 \leq N_{h,2} < 26}$ were assessed. Due to the high number of \glspl{ANN} which needed to be tested, it was not feasible to use all available data but only a subset of it. To find the sufficient number of samples, an \gls{ANN} with two hidden layers of 25 neurons each was tested with varying amounts of input data. The reason for the choice of ${N_{h,1} = N_{h,2} = 25}$ is the fact that the required number of samples increases with the number of degrees of freedom of an \gls{ANN}, which in turn increases with the number of layers as well as with the number of neurons. Therefore, it was reasonable to determine the number of required samples on the largest \gls{ANN} which was going to be tested. As a measure of quality, the minimum cost, i.e. the cross-entropy, achieved by the \gls{ANN} on the test set during 200 epochs of training was used. Due to the influence of the random initialization of the \gls{ANN}, each measurement was repeated for ten different random seeds and the mean of the results was calculated. 

\begin{figure}
	\centering
	\includegraphics[width=\plotwidth]{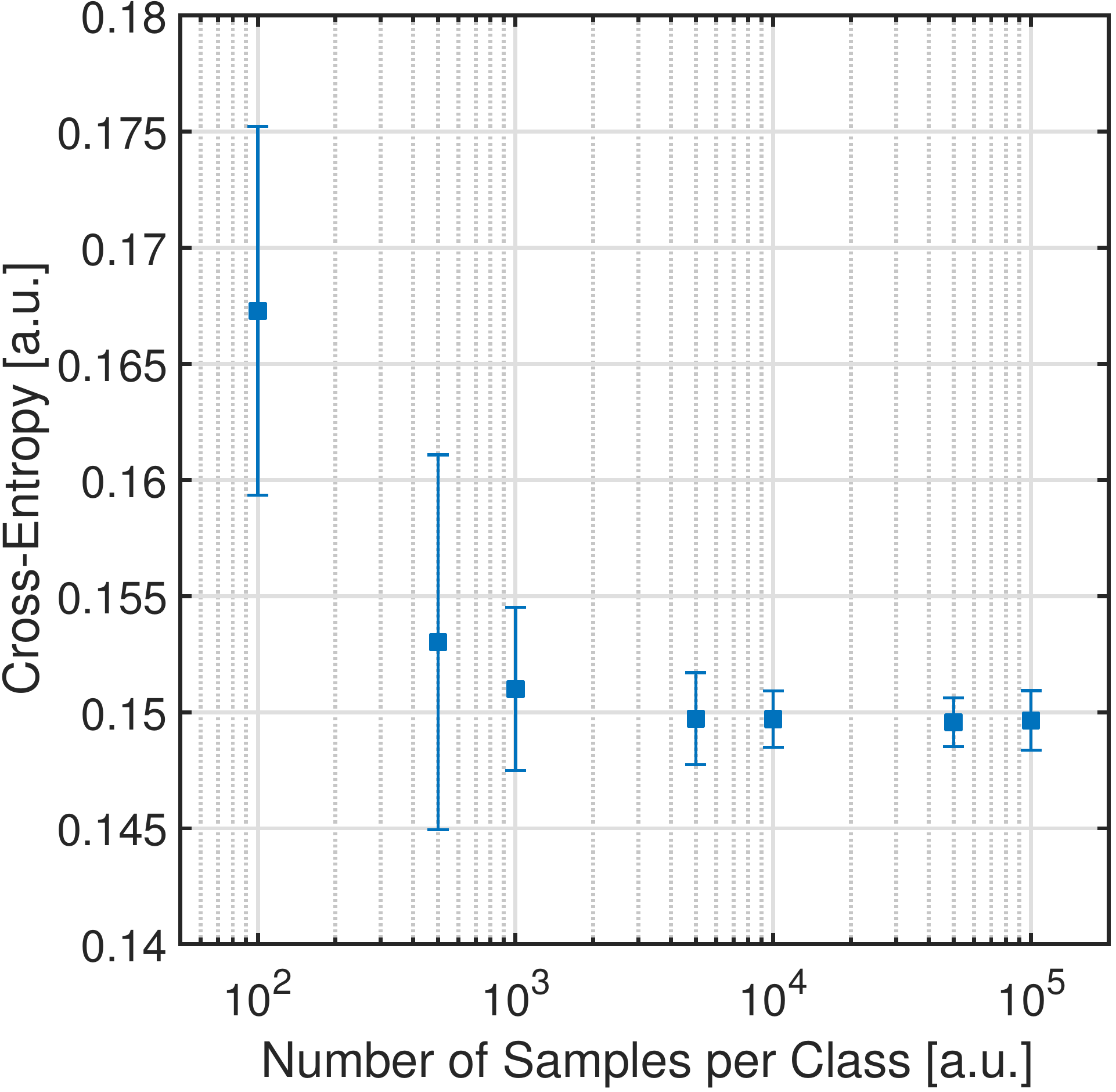}
	\caption{Cross-entropy as a function of the number of samples per class for a ParticleNumberNet with two hidden layers of 25 neurons each. Please note the logarithmic scale of the $x$-axis.}
	\label{fig:SampleNumberVsCrossEntropy}
\end{figure}

Figure \ref{fig:SampleNumberVsCrossEntropy} displays the results of the investigation of the influence of the number of samples on the quality of the \gls{ANN}. The mean and the standard deviation of the cross-entropy decrease with an increasing number of samples but do not improve further in the region above 10,000 samples per class. Therefore, a number of 10,000 samples per class was identified as being sufficient for the investigation of the optimal structure of the ParticleNumberNet.

For the investigation of the optimal structure of the ParticleNumberNet, 10,000 samples per class were shuffled and randomly split into three sets (Table \ref{tab:ParticleNumber_DataDivision}). Subsequently, the lowest cross-entropy achieved by the \gls{ANN} on the test set during 200 epochs of training was determined for each combination of ${1 \leq N_{h,1} < 26}$ and ${0 \leq N_{h,2} < 26}$. Due to the influence of the random initialization of the \glspl{ANN}, each measurement was repeated for ten different random seeds and the mean of the results was calculated. Ultimately, the investigation yielded that no significant improvements could be achieved by the incorporation of a second hidden layer. However, better results were achieved for higher numbers of neurons in the first hidden layer. Therefore, the assessment described above was repeated for ParticleNumberNets with a single hidden layer of up to 100 neurons. The resulting means of the minimum cross-entropies were normalized via a division by the mean of the minimum cross-entropies of the \glspl{ANN} with a single hidden neuron and plotted versus the number of hidden neurons. 

\begin{figure}
	\centering
	\includegraphics[width=\plotwidth]{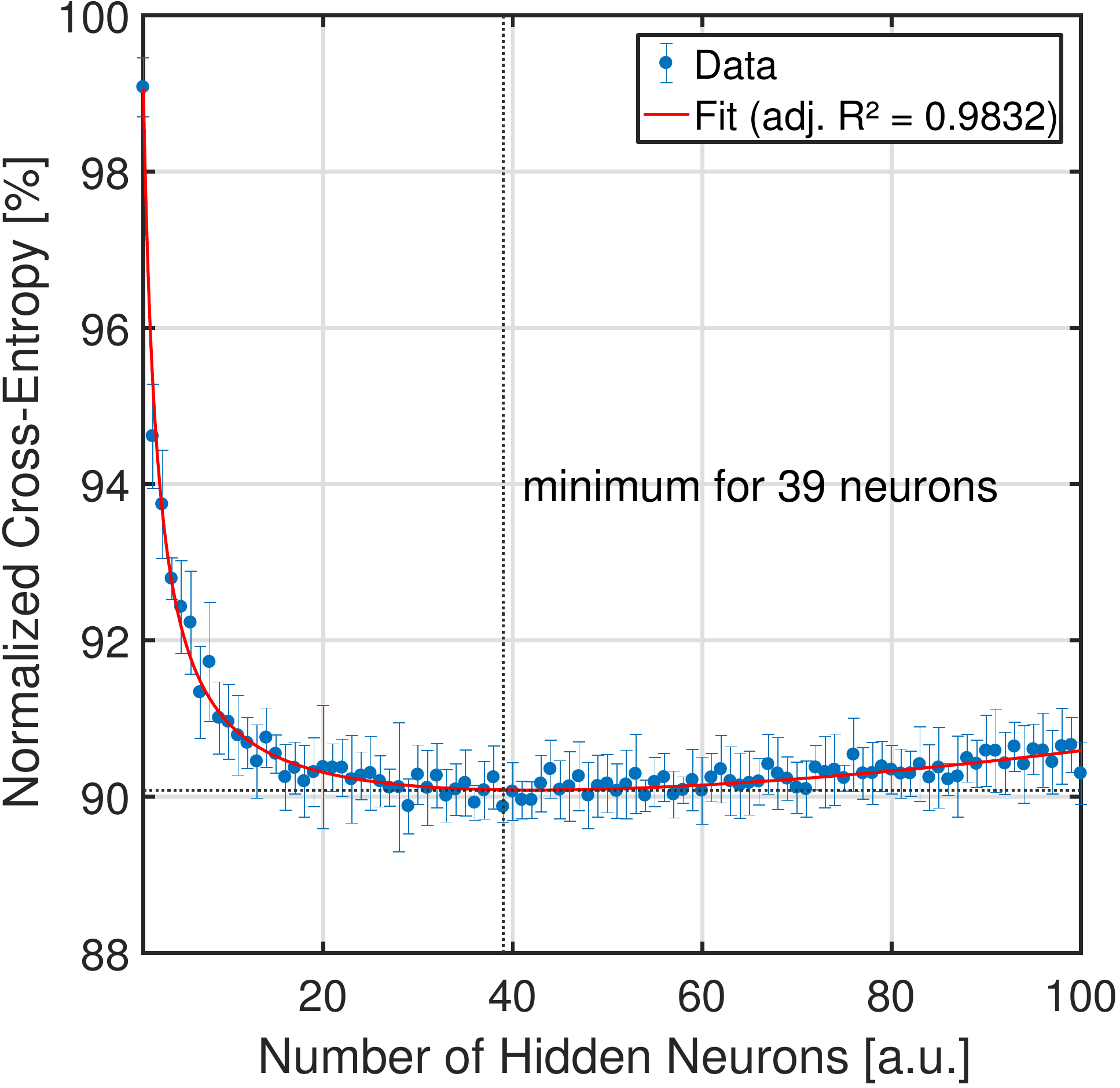}
	\caption{Cross-entropy, normalized based on the maximum cross-entropy which was encountered, for different numbers of neurons in the hidden layer of the ParticleNumberNet.}
	\label{fig:ParticleNumberNet_StructureOptimization_1HiddenLayer}
\end{figure}

The resulting graph (Figure \ref{fig:ParticleNumberNet_StructureOptimization_1HiddenLayer}) was fitted by a function of the form
\begin{equation*}
f(x) = \frac{ax^3 + bx^2 + cx + d}{x+e},
\end{equation*}
which was empirically found to fit the results best and exhibits a minimum for 39 neurons in the hidden layer. It was therefore decided to incorporate 39 hidden neurons in the final structure of the ParticleNumberNet (Table \ref{tab:ParticleNumber_Structure}).

\subsection{Validation and Results}

To validate the proposed method with regard to the classification of the number of primary particles incorporated by an agglomerate, its performance was compared to those of three established automated methods (\gls{WST}, \gls{UE} and \gls{HT}) as well as those of the manual method\footnote{The manual counting of the primary particles was performed with help of the \blockquote{Measure and Label} function of the software \textit{ImageJ} \cite{Schneider.2012}.}, once for synthesized and once for real \gls{TEM} images. To account for the human factor, the manual method was performed independently by two different test persons. 

The three established automated methods were implemented according to the \textit{MATLAB} documentation \cite{MATLAB.2016} and their parameters were empirically optimized, to prevent a discrimination of the established automated methods compared to the proposed method. 

\subsubsection{Synthetic Data}
The set of synthetic \gls{TEM} images for the assessment of the classification performance consisted of 15,000 synthetic \gls{TEM} images for each class of agglomerates, i.e. 90,000 images in total. All samples had different \glspl{PSD}, different transmission coefficients and different degrees of particle deformation. The distortion parameters of the synthesized \gls{TEM} images were chosen randomly according to their distributions, which were determined beforehand based on real \gls{TEM} images.

Due to its extremely labor-intensive nature, the manual method was performed on a reduced set of just 100 images per class, i.e. 600 images in total.

The results of the validation based on synthetic \gls{TEM} images can be found in table \ref{tab:ParticleNumber_Results_synthetic}. The proposed method can outperform the established automated methods but is still inferior to the manual evaluation.

\nColumns=2
\setlength{\firstcolumnwidth}{5 cm}
\setlength{\autocolumnwidth}{(\columnwidth-\tabcolsep*(\nColumns*2)-\firstcolumnwidth)/(\nColumns-1)}

\begin{table}
	\centering
	\caption{Ranking of the mean classification accuracies of the proposed method and established automated methods on a data set of 90,000 synthetic \acrlong{TEM} images, as well as of the manual method on a data set of 600 synthetic \acrlong{TEM} images.}
	\renewcommand{\arraystretch}{1.2}
	\label{tab:ParticleNumber_Results_synthetic}
	\begin{tabular}{>{\centering\bfseries}m{\firstcolumnwidth} >{\centering\arraybackslash}m{\autocolumnwidth}}
		\toprule
		\toprule
		& \textbf{Mean Classification Accuracy} \\
		Test Person A& \SI{88.3}{\percent} \\
		Test Person B& \SI{84.0}{\percent} \\
		Proposed Method& \SI{63.8}{\percent} \\
		Hough Transformation& \SI{55.0}{\percent} \\
		Ultimate Erosion& \SI{48.2}{\percent} \\
		Watershed Transformation& \SI{46.5}{\percent} \\ 
		\bottomrule
	\end{tabular}
\end{table}

\subsubsection{Real Data}
\label{sec:ResultsValidationAndDiscussion-Classification-RealData}
The set of real \gls{TEM} images for the assessments of the classification performance consisted of 500 real \gls{TEM} images of agglomerates. Due to the reason that the actual number of primary particles depicted on a real \gls{TEM} image cannot be determined definitively, the results of the manual measurements of test person A, who performed best on the synthetic \gls{TEM} images (Table \ref{tab:ParticleNumber_Results_synthetic}), were used as ground truth. 

In contrast to the set of synthetic \gls{TEM} images, which was used for the assessment of the classification performance, the set of real \gls{TEM} images did not contain an equal number of samples of each class (Table \ref{tab:NumberOfSamplesPerClass-Classification-Real}), due to the limited availability of suitable \gls{TEM} images. This circumstance should be considered, when trying to compare the results of the assessment of the classification performance based on synthetic \gls{TEM} images to those of the assessment based on real \gls{TEM} images.

The results of the validation based on real \gls{TEM} images can be found in table \ref{tab:ParticleNumber_Results_real}. Just like in the case of synthetic \gls{TEM} images, the proposed method can outperform the three established automated methods but still lacks the accuracy of the manual method.

\nColumns=2
\setlength{\firstcolumnwidth}{5 cm}
\setlength{\autocolumnwidth}{(\columnwidth-\tabcolsep*(\nColumns*2)-\firstcolumnwidth)/(\nColumns-1)}

\begin{table}
	\centering
	\caption{Ranking of the mean classification accuracies of the proposed method, established automated methods and the manual evaluation on a data set of 500 real \acrlong{TEM} images (ground truth: manual evaluation of test person A).}
	\renewcommand{\arraystretch}{1.2}
	\label{tab:ParticleNumber_Results_real}
	\begin{tabular}{>{\centering\bfseries}m{\firstcolumnwidth} >{\centering\arraybackslash}m{\autocolumnwidth}}
		\toprule
		\toprule
		& \textbf{Mean classification accuracy} \\
		Test Person B& \SI{68.2}{\percent} \\
		Proposed Method & \SI{60.8}{\percent} \\
		Ultimate Erosion & \SI{59.0}{\percent} \\
		Watershed Transformation & \SI{53.8}{\percent} \\ 
		Hough Transformation & \SI{23.2}{\percent} \\
		\bottomrule
	\end{tabular}
\end{table}

\section{Primary Particle Area Regression}
The following section will elaborate upon the design, training, validation and results of the \glspl{ANN}, which were used for the regression of the areas of the primary particles of an agglomerate (hereinafter called ParticleAreaNets). As mentioned above, each class of agglomerates, except the mixed sixth class which contained agglomerates with more than five primary particles and was therefore excluded from the analysis, needed to be analyzed by an individual ParticleAreaNet.

\subsection{Training, Validation and Test Data}
The input data used for the training of the ParticleAreaNets was identical to the training data of the ParticleNumberNet but was divided into individual data sets for each class of agglomerates.

The target data sets for the training of the ParticleAreaNets consisted of 100,000 vectors per class, where each of the vectors had a number of components equal to the number of primary particles in an agglomerate of the corresponding class. The components contained the areas of the primary particles.

Just like before, each input and target data set was divided into three subsets (Table \ref{tab:ParticleArea_DataDivision}), to evaluate the generalization abilities of the \glspl{ANN} and to allow the utilization of the early stopping method.

\subsection{Structural Optimization}

The final structures of the ParticleAreaNets and the learning strategies utilized during their training are given in tables \ref{tab:ParticleArea_Structure} and \ref{tab:ParticleArea_NeuronNumber} and in table \ref{tab:ParticleArea_Learning} respectively. 

To yield optimal results, the structures of the ParticleAreaNets were again optimized iteratively. The number of neurons in the input and output layers of the ParticleAreaNets are defined by the number of components of input ($N_i = 13$) and output samples ($1 \leq N_o \leq 5$) respectively. Due to the reasons given before, only \gls{ANN} structures with up to two hidden layers were tested as ParticleAreaNets.

In case of the ParticleAreaNets the rules of thumb (see equations \ref{eq:RuleOfThumb1} and \ref{eq:RuleOfThumb2}) \cite{Heaton.2008}, yield the following numbers of neurons as orientation for the optimization of the hidden layers:

\vspace{-1\abovedisplayskip}
\noindent\begin{tabularx}{\columnwidth}{@{}c@{}X@{}c@{}X@{}c@{}} 
	&
	{\setlength{\mathindent}{0cm}
		\begin{align*} 
		&&1..5 &\leq N_{h,i} < 26
		\end{align*}} &
	 &
	{\setlength{\mathindent}{0cm}
		\begin{align*} 
		&&N_{h,i} &\approx 10..14
		\end{align*}} &
\end{tabularx}
\vspace{-2\belowdisplayskip}

Before the assessment of the optimal number of neurons in the hidden layers of the ParticleAreaNets, the necessary number of input samples was determined. The ParticleAreaNet with the most degrees of freedom, i.e. with the highest number of neurons was used for the evaluation. The ParticleAreaNet with the highest number of neurons is the ParticleAreaNet for agglomerates with five primary particles ($N_o = 5$) and two hidden layers of 25 neurons each (${N_{h,1} = N_{h,2} = 25}$). As a measure of quality of the \gls{ANN}, the minimum cost, i.e. the mean squared error, achieved by the \gls{ANN} on the test set during 200 epochs of training was used. Due to the influence of the random initialization of the \gls{ANN}, each measurement was repeated for ten different random seeds and the mean of the results was calculated.

\begin{figure}
	\centering
	\includegraphics[width=0.94\plotwidth]{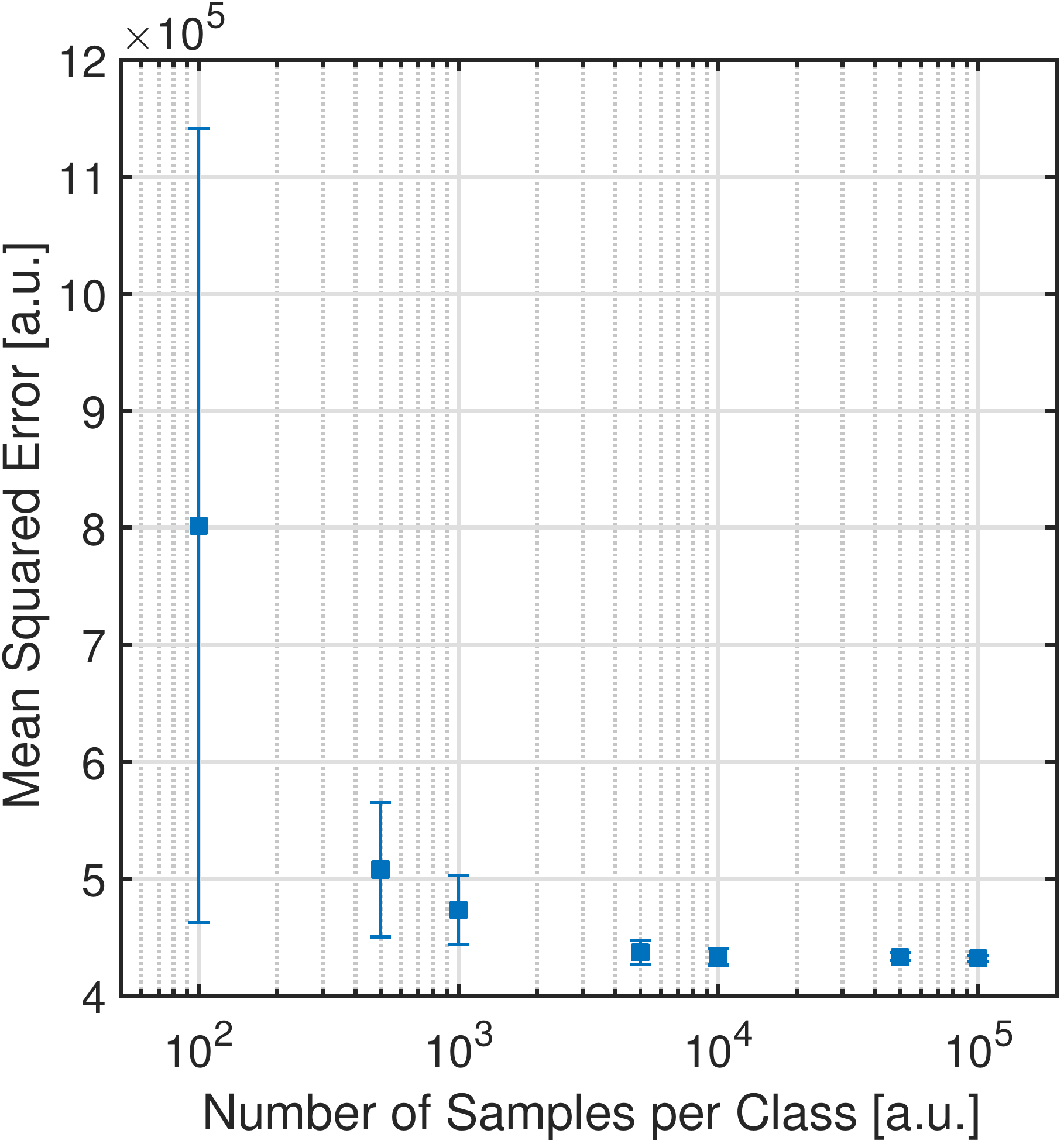}
	\caption{Mean squared error as a function of the number of samples per class for a ParticleAreaNet for agglomerates containing five primary particles with two hidden layers of 25 neurons each. Please note the logarithmic scale of the $x$-axis.}
	\label{fig:SampleNumberVsMSE}
\end{figure}

Figure \ref{fig:SampleNumberVsMSE} displays the results of the investigation of the influence of the number of samples on the quality of the \gls{ANN}. The mean and the standard deviation of the mean squared error decrease with an increasing number of samples, but do not further improve in the region above 10,000 samples per class. Therefore, a number of 10,000 samples per class was identified as being sufficient for the investigation of the optimal structures of the ParticleAreaNets.

\begin{figure}
	\centering
	\includegraphics[width=\plotwidth]{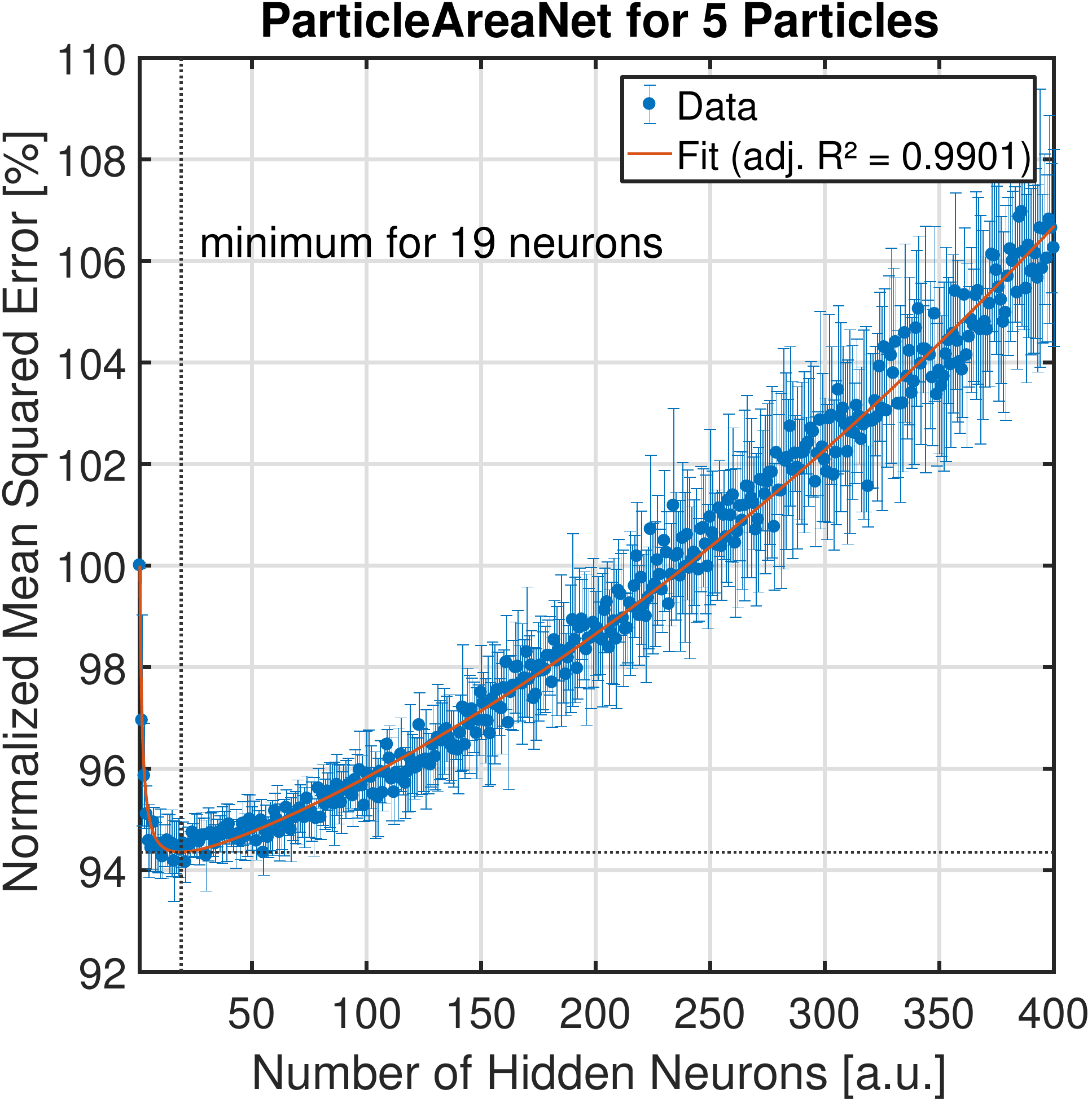}
	\caption{Mean squared error, normalized based on the maximum mean squared error which was encountered, for different numbers of neurons in the hidden layer of the ParticleAreaNet for agglomerates containing five primary particles.}
	\label{fig:ParticleAreaNet5_StructureOptimization_1HiddenLayer}
\end{figure}

For the assessment of the optimal numbers of neurons in the hidden layers of the ParticleAreaNets, all possible combinations of ${1 \leq N_{h,1} < 26}$ and ${0 \leq N_{h,2} < 26}$ were tested for each of the ParticleAreaNets. The assessment was performed similarly to the procedure described in section \ref{sec:PrimaryParticleNumberClassification-StructuralOptimization}. For each of the ParticleAreaNets, 10,000 samples were shuffled and randomly split into three sets (Table \ref{tab:ParticleArea_DataDivision}). Subsequently, the lowest mean squared errors achieved by the ParticleAreaNets on the respective test sets during 200 epochs of training were determined for each combination of ${1 \leq N_{h,1} < 26}$ and ${0 \leq N_{h,2} < 26}$. Due to the influence of the random initialization of the \glspl{ANN}, each measurement was repeated for ten different random seeds and the means of the results were calculated for the different ParticleAreaNets. Ultimately, the investigation yielded that no significant improvements could be achieved by the incorporation of a second hidden layer. However, better results were achieved for higher numbers of neurons in the first hidden layer. Therefore, the assessment described above was repeated for ParticleAreaNets with single hidden layers of up to 400 neurons. The resulting means of the minimum mean squared errors were normalized via a division by the mean of the minimum mean squared errors of the \glspl{ANN} with a single hidden neuron and plotted versus the number of hidden neurons. The resulting graphs (e.g. figure \ref{fig:ParticleAreaNet5_StructureOptimization_1HiddenLayer}; for further information please refer to the supplementary material) were fitted by functions of the form
\begin{equation*}
f(x) = \frac{ax^3 + bx^2 + cx + d}{x+e},
\end{equation*}
which were empirically found to fit the results best. The numbers of neurons in the hidden layers of the final structures of the ParticleAreaNets (Tables \ref{tab:ParticleArea_Structure} and \ref{tab:ParticleArea_NeuronNumber}) were set to the number of neurons where the respective fits exhibited their minima.

\subsection{Validation and Results}

To validate the proposed method with regard to the measurement of \glspl{PSD}, its performance was compared to those of the three established automated methods as well as those of the manual method\footnote{The manual measurements of the areas of the primary particles were performed with help of the \blockquote{Elliptical selections} and the \blockquote{Measure and Label} function of the software \textit{ImageJ} \cite{Schneider.2012}.}, once for synthesized and once for real \gls{TEM} images. To account for the human factor, the manual method was performed independently by two different test persons.

\subsubsection{Synthetic Data}
The set of samples of synthetic \gls{TEM} images for the assessments of the performance of the \gls{PSD} measurement consisted of 300 samples with 500 agglomerates each. The number of agglomerates was chosen so that the minimum number of particles of a sample was 500 (for a sample containing only single particles), which is considered to be the necessary minimum for a good statistical quality of \gls{PSD} measurements \cite{Carpenter.1998}. All samples had different \glspl{GMD} and \glspl{GSD}, different distributions of primary particles per agglomerate, different transmission coefficients and different degrees of particle deformation. The distortion parameters of the synthesized \gls{TEM} images were chosen randomly according to their distributions, which were determined beforehand based on real \gls{TEM} images.

Due to its extremely labor-intensive nature, the manual method was performed on a reduced set of just one sample of 500 agglomerates.

For the assessment of the \gls{PSD} measurement performance, the output \gls{GMD} $d_{g,O}$ and output \gls{GSD} $\sigma_{g,O}$ of each sample were determined based on the output primary particle areas, which were measured via the different methods and compared to the corresponding target \gls{GMD} $d_{g,T}$ and target \gls{GSD} $\sigma_{g,T}$, which were determined based on the known target primary particle areas. Subsequently, the relative errors $E_{d_g}$ and $E_{\sigma_g}$ of the \gls{GMD} and \gls{GSD} were calculated: 

\vspace{-1.5\abovedisplayskip}
\noindent\begin{tabularx}{\columnwidth}{@{}c@{}X@{}c@{}X@{}c@{}} 
	 &
	{\setlength{\mathindent}{0cm}
		\begin{align} 
		\label{eq:ErrorGMD}
		&&E_{d_g}&= \frac{d_{g,O}-d_{g,T}}{d_{g,T}}
		\end{align}} &
	 &
	{\setlength{\mathindent}{0cm}
		\begin{align} 
		\label{eq:ErrorGSD}
		&&E_{\sigma_g}&= \frac{\sigma_{g,O}-\sigma_{g,T}}{\sigma_{g,T}}
		\end{align}} &
\end{tabularx}
\vspace{-1.5\belowdisplayskip}

Ultimately, the relative errors of the \gls{GMD} and \gls{GSD} were plotted in form of histograms for each of the methods.

\begin{figure}
	\centering
	\subfiguretitle{Proposed Method}
	\includegraphics[width=\plotwidth]{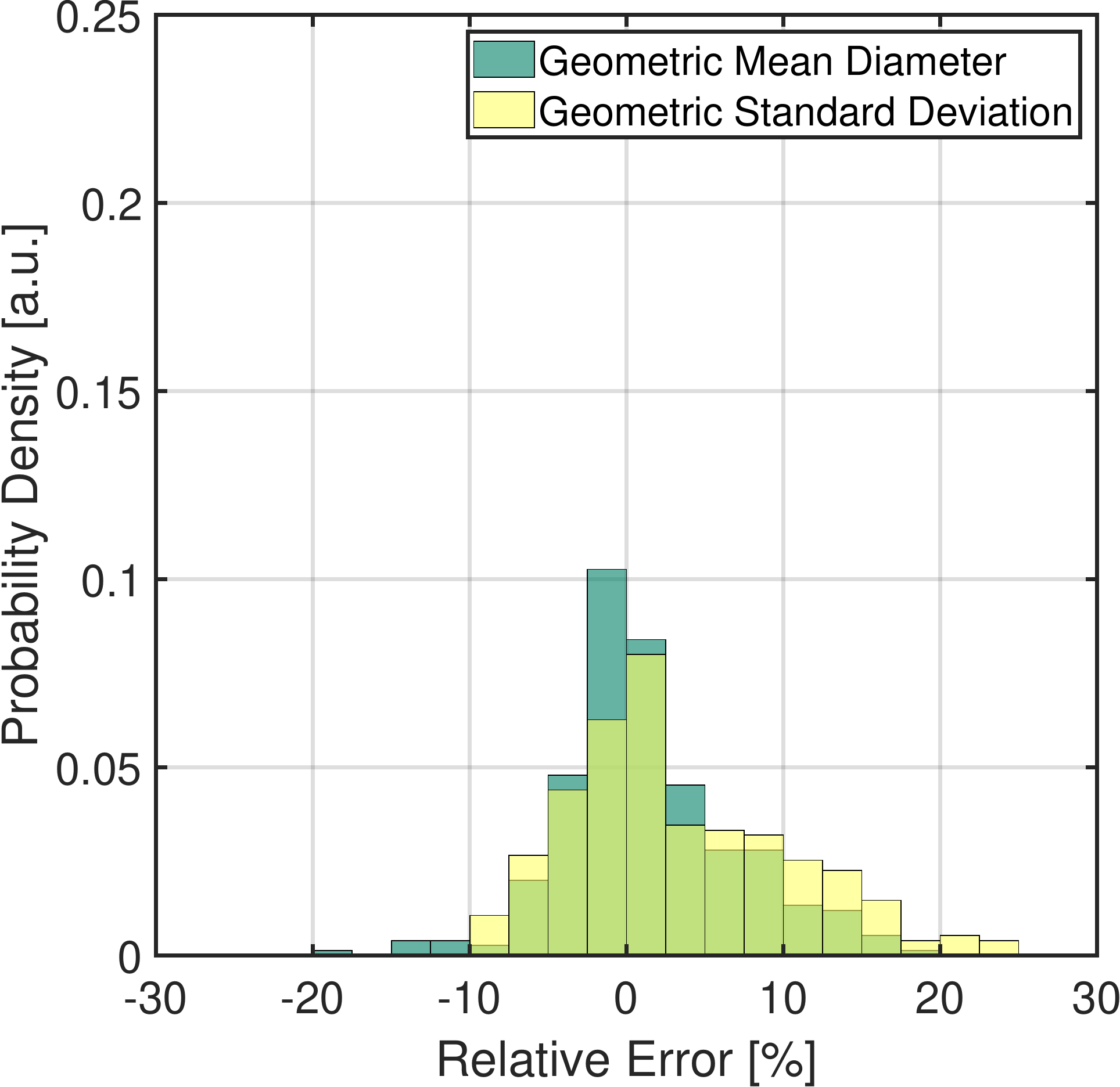}
	\caption{Histograms of the errors of \acrlong{GMD} and \acrlong{GSD} determined by the proposed method when being applied to synthetic \acrlong{TEM} images.}
	\label{fig:PSD_Synthetic_ANN}
\end{figure}

Figure \ref{fig:PSD_Synthetic_ANN} exemplarily depicts the histograms of the relative errors of the \gls{GMD} and \gls{GSD} of the primary \glspl{PSD} of 300 samples of 500 synthetic \gls{TEM} images each, which were determined via the proposed method. It can be noticed that the proposed method tends to overestimate the \gls{GMD} as well as the \gls{GSD} of primary \glspl{PSD}. For the results of the other methods, which were tested, please refer to the supplementary material.

\subsubsection{Real Data}
The set of real \gls{TEM} images for the assessments of the performance of the \gls{PSD} measurement consisted of 500 real \gls{TEM} images in total. Due to the reason that the actual areas of the primary particles depicted on a real \gls{TEM} image cannot be determined definitively, the results of the manual measurements of test person A, who performed best on the synthetic data, were used as ground truth to calculate the relative errors of the \gls{GMD} and the \gls{GSD} according to equation \ref{eq:ErrorGMD} and \ref{eq:ErrorGSD} respectively. 

It is hardly possible to evaluate the performances of the tested methods concerning the measurement of \glspl{PSD} of real samples based on just one sample. It can, however, help to get a basic idea of the potential of the methods.

The deviations of the \gls{GMD} and the \gls{GSD}, which were determined during the validation can be found in table \ref{tab:ParticleArea_Results}. The proposed method clearly outperforms the three established methods with respect to the determination of the \gls{GMD}. In case of the determination of the \gls{GSD}, only the \gls{UE} can outperform the proposed method. In general, all automated methods lack the accuracy of the manual method.

\nColumns=3
\setlength{\firstcolumnwidth}{3 cm}
\setlength{\autocolumnwidth}{(\columnwidth-\tabcolsep*(\nColumns*2)-\firstcolumnwidth)/(\nColumns-1)}

\begin{table}
	\centering
	\caption{Relative errors of the \acrlong{GMD} and the \acrlong{GSD} determined via the proposed method, established automated methods and the manual evaluation on a data set of 500 real \acrlong{TEM} images (ground truth: manual evaluation of test person A).}
	\renewcommand{\arraystretch}{1.2}
	\label{tab:ParticleArea_Results}
	\begin{tabular}{>{\centering\bfseries}m{\firstcolumnwidth} >{\centering\bfseries}m{\autocolumnwidth} >{\centering\arraybackslash}m{\autocolumnwidth}}
		\toprule
		\toprule
		& \textbf{Relative Error of the \gls{GMD}} & \textbf{Relative Error of the \gls{GSD}} \\
		Proposed Method & \SI{4.1}{\percent} & \SI{5.1}{\percent} \\
		Watershed Transformation& \SI{-11.6}{\percent} & \SI{6.1}{\percent} \\
		Ultimate Erosion & \SI{-10.7}{\percent} & \SI{1.5}{\percent} \\
		Hough Transformation & \SI{-20.1}{\percent} & \SI{8.7}{\percent} \\
		Test Person B & \SI{-4.3}{\percent} & \SI{-1.0}{\percent} \\	
		\bottomrule
	\end{tabular}
\end{table}

\section{Conclusion}
Within this work, a novel method for the analysis of the size of primary particles of agglomerates on \gls{TEM} images was successfully implemented and validated. Therefore, a two-stage strategy was applied: During the first stage, each agglomerate is assigned to a class, depending on the number of primary particles it incorporates, by a classification \gls{ANN}. Subsequently, the areas of the incorporated primary particles are analyzed by a regression \gls{ANN} that is specific to the class of the agglomerate. 

Regarding the classification performance, the proposed method is superior to the \gls{WST}, \gls{HT} and \gls{UE}. However, for reasons of fairness, it should be noted that none of the established automated methods was designed specifically for the determination of the number of primary particles in agglomerates. Another result of the evaluation of the classification performance is that the proposed method, as well as the established automated methods, are clearly outperformed by the manual method.

With respect to the determination of primary \glspl{PSD} of agglomerates on \gls{TEM} images, the evaluation of the performance is especially difficult because only a single real sample of \gls{TEM} images was available. However, for this sample, as well as for samples of synthetic \gls{TEM} images, the validation yielded very promising results for the proposed method. Again, the manual method could outperform all other tested methods.

All in all, the proposed method is not able to match the accuracy of the manual evaluation. Nevertheless, it is at least able to compete with established automated methods, which have been used and improved for several decades. It may be assumed that its performance could be further improved and may therefore be able to outperform the established automated methods more clearly and perhaps even reach the accuracy of the manual method while offering significantly shorter measurement times. 

\section*{Acknowledgment}

The authors acknowledge the support by the Deutsche Forschungsgemeinschaft (DFG) in scope of the research group 2284 "Model-based scalable gas-phase synthesis of complex nanoparticles". All authors declare that they have no competing interests.

\appendix

\bibliographystyle{elsarticle-num}
\bibliography{bibliography}

\floatplacement{table}{H}

\section{Preprocessing}
\label{app:Preprocessing}
\vspace{-20pt}
\nColumns=2
\setlength{\firstcolumnwidth}{2 cm}
\setlength{\autocolumnwidth}{(\columnwidth-\tabcolsep*(\nColumns*2)-\firstcolumnwidth)/(\nColumns-1)}

\begin{table}
	\centering
	\caption{Agglomerate features determined during the preprocessing process.}
	\renewcommand{\arraystretch}{1.2}
	\label{tab:AgglomerateFeatures}
	\begin{tabular}{>{\centering\bfseries}m{\firstcolumnwidth} >{\arraybackslash}m{\autocolumnwidth}}
		\toprule
		\toprule
		Area & Number of pixels within the agglomerate.\\
		Convex area & Number of pixels within the convex hull of the agglomerate.\\
		Eccentricity & Eccentricity of an ellipse with second moments equal to those of the agglomerate. \\
		Equivalent diameter & Diameter of a circle with an area equal to the area of the agglomerate.\\
		Extent & Ratio of the areas of the agglomerate and its bounding box.\\
		Filled area & Number of pixels within the agglomerate, after all potential holes have been filled.\\
		Major axis length & Length of the major axis of an ellipse with normalized second central moments equal to those of the agglomerate.\\
		Minor axis length & Length of the minor axis of an ellipse with normalized second central moments equal to those of the agglomerate.\\
		Perimeter & Number of pixels surrounding the agglomerate.\\
		Solidity & Ratio of area and convex area of the agglomerate.\\
		Minimum intensity & Intensity of the darkest pixel in the region.\\
		Maximum intensity & Brightest pixel in the region.\\
		Mean intensity & Mean of the intensities of the pixels of the region.\\
		\bottomrule
	\end{tabular}
\end{table}

\section{Primary Particle Number Classification}
\label{app:ParticleNumberNet}

\vspace{-20pt}

\nColumns=4
\setlength{\firstcolumnwidth}{2.7 cm}
\setlength{\autocolumnwidth}{(\columnwidth-\tabcolsep*(\nColumns*2)-\firstcolumnwidth)/(\nColumns-1)}

\begin{table}
	\centering
	\caption{Structural features of the \acrlong{ANN} for the classification of agglomerates with respect to the number of primary particles.}
	\renewcommand{\arraystretch}{1.2}
	\label{tab:ParticleNumber_Structure}
	\begin{tabular}{>{\centering\bfseries}m{\firstcolumnwidth} >{\centering\arraybackslash}m{\autocolumnwidth} >{\centering\arraybackslash}m{\autocolumnwidth} >{\centering\arraybackslash}m{\autocolumnwidth}}
		\toprule
		\toprule
		& \textbf{Input layer} & \textbf{Hidden layer} & \textbf{Output layer}\\
		Propagation function & identity & weighted sum & weighted sum \\
		Activation function & identity & hyperbolic tangent & softmax \\
		Output function & identity & identity & identity \\
		Number of neurons & 13 & 39 & 6 \\
		\bottomrule
	\end{tabular}
\end{table}

\vspace{-20pt}

\nColumns=2
\setlength{\firstcolumnwidth}{4.4 cm}
\setlength{\autocolumnwidth}{(\columnwidth-\tabcolsep*(\nColumns*2)-\firstcolumnwidth)/(\nColumns-1)}

\begin{table}
	\centering
	\caption{Learning features of the \acrlong{ANN} for the classification of agglomerates with respect to the number of primary particles.}
	\renewcommand{\arraystretch}{1.2}
	\label{tab:ParticleNumber_Learning}
	\begin{tabular}{>{\centering\bfseries}m{\firstcolumnwidth} >{\centering\arraybackslash}m{\autocolumnwidth}}
		\toprule
		\toprule
		Cost function & cross-entropy \\
		Learning rule & scaled conjugate gradient backpropagation \cite{Moller.1993}\\
		Overfitting prevention & early stopping \\
		Number of samples per class & 100,000 \\
		Number of epochs & 100,000\\
		\bottomrule
	\end{tabular}
\end{table}

\vspace{-20pt}

\nColumns=2
\setlength{\firstcolumnwidth}{4.4 cm}
\setlength{\autocolumnwidth}{(\columnwidth-\tabcolsep*(\nColumns*2)-\firstcolumnwidth)/(\nColumns-1)}

\begin{table}
	\centering
	\caption{Division of the data used to train and evaluate the \acrlong{ANN} for the classification of agglomerates with respect to the number of primary particles.}
	\renewcommand{\arraystretch}{1.2}
	\label{tab:ParticleNumber_DataDivision}
	\begin{tabular}{>{\centering\bfseries}m{\firstcolumnwidth} >{\centering\arraybackslash}m{\autocolumnwidth}}
		\toprule
		\toprule
		Training set: & \SI{70}{\percent} \\
		Validation set: & \SI{15}{\percent} \\ 
		Test set: & \SI{15}{\percent} \\
		\bottomrule
	\end{tabular}
\end{table}

\vspace{-20pt}

\nColumns=2
\setlength{\firstcolumnwidth}{4.4 cm}
\setlength{\autocolumnwidth}{(\columnwidth-\tabcolsep*(\nColumns*2)-\firstcolumnwidth)/(\nColumns-1)}

\begin{table}
	\centering
	\caption{Composition of the set of real \acrlong{TEM} images for the assessment of the primary particle number classification performance as classified by test person A.}
	\label{tab:NumberOfSamplesPerClass-Classification-Real}
	\renewcommand{\arraystretch}{1.2}
	\begin{tabular}{>{\centering\bfseries}m{\firstcolumnwidth} >{\centering\arraybackslash}m{\autocolumnwidth}}
		\toprule
		\toprule
		& \textbf{Number of samples} \\
		1 particle&234\\
		2 particles&153\\
		3 particles&64\\
		4 particles&26\\
		5 particles&9\\
		5+ particles&14\\
		\bottomrule
	\end{tabular}
\end{table}

\section{Primary Particle Area Regression}
\label{app:ParticleAreaNets}

\vspace{-20pt}

\nColumns=4
\setlength{\firstcolumnwidth}{2.7 cm}
\setlength{\autocolumnwidth}{(\columnwidth-\tabcolsep*(\nColumns*2)-\firstcolumnwidth)/(\nColumns-1)}

\begin{table}
	\centering
	\caption{Structural features of the \acrlongpl{ANN} for the regression of primary particle areas.}
	\renewcommand{\arraystretch}{1.2}
	\label{tab:ParticleArea_Structure}
	\begin{tabular}{>{\centering\bfseries}m{\firstcolumnwidth} >{\centering\arraybackslash}m{\autocolumnwidth} >{\centering\arraybackslash}m{\autocolumnwidth} >{\centering\arraybackslash}m{\autocolumnwidth}}
		\toprule
		\toprule
		& \textbf{Input layer} & \textbf{Hidden layer} & \textbf{Output layer}\\
		Propagation function & identity & weighted sum & weighted sum \\
		Activation function & identity & hyperbolic tangent & identity \\
		Output function & identity & identity & identity \\
		\bottomrule
	\end{tabular}
\end{table}

\vspace{-20pt}

\nColumns=4
\setlength{\firstcolumnwidth}{3 cm}
\setlength{\autocolumnwidth}{(\columnwidth-\tabcolsep*(\nColumns*2)-\firstcolumnwidth)/(\nColumns-1)}

\begin{table}
	\centering
	\caption{Numbers of neurons in the layers of the \acrlongpl{ANN} for the regression of primary particle areas.}
	\renewcommand{\arraystretch}{1.2}
	\label{tab:ParticleArea_NeuronNumber}
	\begin{tabular}{>{\centering\bfseries}m{\firstcolumnwidth} >{\centering\arraybackslash}m{\autocolumnwidth} >{\centering\arraybackslash}m{\autocolumnwidth} >{\centering\arraybackslash}m{\autocolumnwidth}}
		\toprule
		\toprule
		& \multicolumn{3}{c}{\textbf{Number of neurons}}\\
		\textbf{Number of primary particles} & \textit{Input layer} & \textit{Hidden layer} & \textit{Output layer}\\
		1 & 13 & 11 & 1 \\
		2 & 13 & 124 & 2 \\
		3 & 13 & 104 & 3 \\
		4 & 13 & 29 & 4 \\
		5 & 13 & 19 & 5 \\
		\bottomrule
	\end{tabular}
\end{table}

\vspace{-20pt}

\nColumns=2
\setlength{\firstcolumnwidth}{4.4 cm}
\setlength{\autocolumnwidth}{(\columnwidth-\tabcolsep*(\nColumns*2)-\firstcolumnwidth)/(\nColumns-1)}

\begin{table}
	\centering
	\caption{Division of the data used to train and evaluate the \acrlongpl{ANN} for the regression of primary particle areas.}
	\renewcommand{\arraystretch}{1.2}
	\label{tab:ParticleArea_DataDivision}
	\begin{tabular}{>{\centering\bfseries}m{\firstcolumnwidth} >{\centering\arraybackslash}m{\autocolumnwidth}}
		\toprule
		\toprule
		Training set: & \SI{70}{\percent} \\
		Validation set: & \SI{15}{\percent} \\ 
		Test set: & \SI{15}{\percent} \\
		\bottomrule
	\end{tabular}
\end{table}

\vspace{-20pt}

\nColumns=2
\setlength{\firstcolumnwidth}{4.4 cm}
\setlength{\autocolumnwidth}{(\columnwidth-\tabcolsep*(\nColumns*2)-\firstcolumnwidth)/(\nColumns-1)}

\begin{table}
	\centering
	\caption{Learning features of the \acrlongpl{ANN} for the regression of primary particle areas.}
	\renewcommand{\arraystretch}{1.2}
	\label{tab:ParticleArea_Learning}
	\begin{tabular}{>{\centering\bfseries}m{\firstcolumnwidth} >{\centering\arraybackslash}m{\autocolumnwidth}}
		\toprule
		\toprule
		Cost function & mean squared error \\
		Learning rule & scaled conjugate gradient backpropagation \cite{Moller.1993}\\
		Overfitting prevention & early stopping \\
		Number of samples & 100,000 \\
		Number of epochs & 100,000\\
		\bottomrule
	\end{tabular}
\end{table}

\end{document}